\documentclass{article}

\usepackage{times}
\usepackage{graphicx} 
\usepackage{subfigure}

\usepackage{natbib}
\usepackage{floatrow}

\usepackage{hyperref}

\usepackage[accepted]{icml2016}

\usepackage{hyperref}
\usepackage{url}
\usepackage{amsmath}
\usepackage{amsthm}
\usepackage{amssymb}
\usepackage{graphicx}
\usepackage{colortbl}
\usepackage{subfigure}
\usepackage{mathtools}
\usepackage{multirow}
\usepackage{color}
\usepackage{soul}
\usepackage{adjustbox}
\usepackage{booktabs}

\DeclareCaptionLabelFormat{andtable}{#1~#2  \&  \tablename~\thetable}

\usepackage{changepage}
\newcommand{\vect}[1]{\mathbf{#1}}
\newcommand{\D}{\mathcal{D}}
\newcommand{\x}{\vect{x}}

\newcommand{\M}{\vect{M}}
\newcommand{\R}{\mathbb{R}}

\newcommand{\fc}{f_{c}}
\newcommand{\hc}{h_{c}}
\newcommand{\hf}{h_{f}}
\newcommand{\ff}{f_{f}}
\newcommand{\gtop}{g}
\newcommand{\xsi}{\x^{(i)}}
\newcommand{\ysi}{y^{(i)}}

\newcommand{\bmx}[0]{\begin{bmatrix}}
  \newcommand{\emx}[0]{\end{bmatrix}}

\renewcommand{\cite}[1]{\citep{#1}}
\newfloatcommand{capbtabbox}{table}[][\FBwidth]

\icmltitlerunning{Dynamic Capacity Networks}

\begin{document}

\twocolumn[ \icmltitle{Dynamic Capacity Networks}


\icmlauthor{Amjad Almahairi$^{\ast}$}{amjad.almahairi@umontreal.ca}
\icmlauthor{Nicolas Ballas$^{\ast}$}{nicolas.ballas@umontreal.ca}
\icmlauthor{Tim Cooijmans$^{\ast}$}{tim.cooijmans@umontreal.ca}
\icmlauthor{Yin Zheng$^{\dagger}$}{yin.zheng@hulu.com}
\icmlauthor{Hugo Larochelle$^{\star}$}{hlarochelle@twitter.com}
\icmlauthor{Aaron Courville$^{\ast}$}{aaron.courville@umontreal.ca}
\icmladdress{$^{\ast}$MILA, Universit\'e de Montr\'eal, Qu\'ebec, Canada \\
  $^{\dagger}$Hulu LLC. Beijing, China \\
  $^{\star}$Twitter, Cambridge, MA, USA}


\icmlkeywords{conditional computation, attention, dynamic capacity}

\vskip 0.3in ]

\begin{abstract}
  We introduce the Dynamic Capacity Network (DCN), a neural network
  that can adaptively assign its capacity across different portions of
  the input data.  This is achieved by combining modules of two types:
  low-capacity sub-networks and high-capacity sub-networks.  The
  low-capacity sub-networks are applied across most of the input, but
  also provide a guide to select a few portions of the input on which
  to apply the high-capacity sub-networks. The selection is made using
  a novel gradient-based attention mechanism, that efficiently
  identifies input regions for which the DCN's output is most
  sensitive and to which we should devote more capacity.  We focus our
  empirical evaluation on the Cluttered MNIST and SVHN image datasets.
  Our findings indicate that DCNs are able to drastically reduce the
  number of computations, compared to traditional convolutional neural
  networks, while maintaining similar or even better performance.
\end{abstract}

\section{Introduction}

Deep neural networks have recently exhibited state-of-the-art
performance across a wide range of tasks, including object
recognition~\cite{szegedy2014going} and speech
recognition~\cite{graves2014towards}. Top-performing systems, however,
are based on very deep and wide networks that are computationally
intensive. One underlying assumption of many deep models is that all
input regions contain the same amount of information.  Indeed,
convolutional neural networks apply the same set of filters uniformly
across the spatial input~\cite{szegedy2014going}, while recurrent
neural networks apply the same transformation at every time
step~\cite{graves2014towards}.  Those networks lead to time-consuming
training and inference (prediction), in large part because they
require a large number of weight/activation multiplications.

Task-relevant information, however, is often not uniformly distributed
across input data. For example, objects in images are spatially
localized, i.e.\ they exist only in specific regions of the image.
This observation has been exploited in \emph{attention-based} systems
\cite{mnih2014recurrent}, which can reduce computations significantly
by learning to selectively focus or ``attend'' to few, task-relevant,
input regions. Attention employed in such systems is often referred to
as ``hard-attention'', as opposed to ``soft-attention'' which
integrates smoothly all input regions.  Models of hard-attention
proposed so far, however, require defining an explicit predictive
model, whose training can pose challenges due to its
non-differentiable cost.

In this work we introduce the \emph{Dynamic Capacity Network} (DCN)
that can adaptively assign its capacity across different portions of
the input, using a gradient-based hard-attention process.
The DCN combines two types of modules: small, low-capacity,
sub-networks, and large, high-capacity, sub-networks.  The
low-capacity sub-networks are active on the whole input, but are used
to direct the high-capacity sub-networks, via our attention mechanism,
to task-relevant regions of the input.

A key property of the DCN's hard-attention mechanism is that it
\emph{does not} require a policy network trained by reinforcement
learning.  Instead, we can train DCNs end-to-end with backpropagation.
We evaluate a DCN model on the attention benchmark task Cluttered
MNIST~\cite{mnih2014recurrent}, and show that it outperforms the state
of the art.

In addition, we show that the DCN's attention mechanism can deal with
situations where it is difficult to learn a task-specific attention
policy due to the lack of appropriate data.
This is often the case when training data is mostly canonicalized,
while at test-time the system is effectively required to perform
transfer learning and deal with substantially different, noisy
real-world images. The Street View House Numbers (SVHN) dataset
~\cite{netzer2011reading} is an example of such a dataset. The task
here is to recognize multi-digit sequences from real-world pictures of
house fronts; however, most digit sequences in training images are
well-centred and tightly cropped, while digit sequences of test images
are surrounded by large and cluttered backgrounds.  Learning an
attention policy that focuses only on a small portion of the input can
be challenging in this case, unless test images are pre-processed to
deal with this discrepancy~\footnote{This is the common practice in
  previous work on this dataset, e.g. \cite{goodfellow2013multi,
    ba2014multiple, jaderberg2015spatial}}.  DCNs, on the other hand,
can be leveraged in such transfer learning scenarios, where we learn
low and high capacity modules independently and only combine them
using our attention mechanism at test-time.  In particular, we show
that a DCN model is able to efficiently recognize multi-digit
sequences, directly from the original images, without using any prior
information on the location of the digits.

Finally, we show that DCNs can perform efficient region selection, in
both Cluttered MNIST and SVHN, which leads to significant
computational advantages over standard convolutional models.

\section{Dynamic Capacity Networks}




In this section, we describe the Dynamic Capacity Network (DCN) that
dynamically distributes its network capacity across an input.

We consider a deep neural network $h$, which we decompose into two
parts: $h(\x) = \gtop(f(\x))$ where $f$ and $\gtop$ represent
respectively the \emph{bottom layers} and \emph{top layers} of the
network $h$ while $\x$ is some input data. Bottom layers $f$ operate
directly on the input and output a representation, which is composed
of a collection of vectors each of which represents a region in the
input.  For example, $f$ can output a feature map, i.e.\ vectors of
features each with a specific spatial location, or a probability map
outputting probability distributions at each different spatial
location.  Top layers $\gtop$ consider as input the bottom layers'
representations $f(x)$ and output a distribution over labels.


DCN introduces the use of two alternative sub-networks for the bottom
layers $f$: the \emph{coarse layers} $\fc$ or the \emph{fine layers}
$\ff$, which differ in their capacity.  The fine layers correspond to
a high-capacity sub-network which has a high-computational
requirement, while the coarse layers constitute a low-capacity
sub-network.  Consider applying the top layers only on the fine
representation, i.e.\ $\hf(\x) = \gtop(\ff(\x))$.  We refer to the
composition $\hf = \gtop \circ \ff$ as the \emph{fine model}.  We
assume that the fine model can achieve very good performance, but is
computationally expensive. Alternatively, consider applying the top
layers only on the coarse representation, i.e.
$\hc(\x) = \gtop(\fc(\x))$. We refer to this composition
$\hc = \gtop \circ \fc$ as the \emph{coarse model}. Conceptually, the
coarse model can be much more computationally efficient, but is
expected to have worse performance than the fine model.

The key idea behind DCN is to have $\gtop$ use representations from
either the coarse or fine layers in an adaptive, dynamic way.
Specifically, we apply the coarse layers $\fc$ on the whole input
$\x$, and leverage the fine layers $\ff$ only at a few ``important''
input regions.
This way, the DCN can leverage the capacity of $\ff$, but at a lower
computational cost, by applying the fine layers only on a small
portion of the input.  To achieve this, DCN requires an attentional
mechanism, whose task is to identify good input locations on which to
apply $\ff$.
In the remainder of this section, we focus on 2-dimensional inputs.
However, our DCN model can be easily extended to be applied to any
type of N-dimensional data.

\subsection{Attention-based Inference}
\label{sec:prediction}

\begin{figure}[t]
  \center
  \includegraphics[width=0.9\columnwidth]{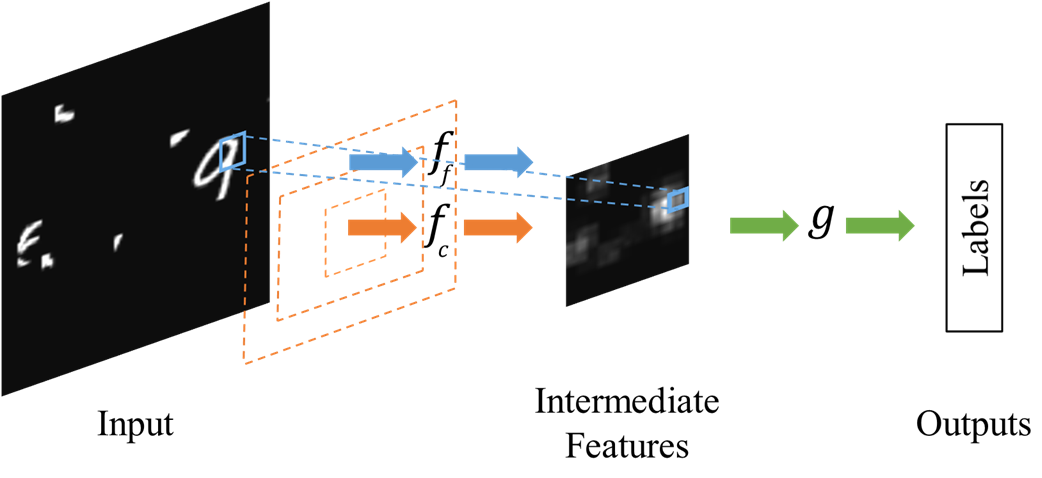}
  \caption{DCN overview. Our model applies the coarse layers on the
    whole image to get $\fc(\x)$, chooses a set of salient patches
    $\vect{X}^{s}$, applies the fine layers only on the salient
    patches $\vect{X}^{s}$ to obtain a set of few fine representation
    vectors $\ff(\vect{X}^{s})$, and finally combines them to make its
    prediction.}
  \label{figure:DCN}
\end{figure}

In DCN, we would like to obtain better predictions than those made by
the coarse model $\fc$ while keeping the computational requirement
reasonable.  This can be done by selecting a few \emph{salient} input
regions on which we use the fine representations instead of the coarse
ones.
DCN inference therefore needs to identify the important regions in the
input with respect to the task at hand.  For this, we use a novel
approach for attention that uses backpropagation in the coarse model
$\hc$ to identify \emph{few vectors in the coarse representation} to
which the distribution over the class label is most sensitive. These
vectors correspond to input regions which we identify as
\emph{salient} or task-relevant.



Given an input image $\x$, we first apply the coarse layers on all
input regions to compute the coarse representation vectors:
\begin{equation}
  \fc(\x) = \{ \vect{c}_{i,j} \mid (i,j) \in \left[1, s_1\right] \times
  \left[1, s_2\right] \},
  \label{eq:coarse}
\end{equation}
where $s_{1}$ and $s_{2}$ are spatial dimensions that depend on the
image size and $\vect{c}_{i,j} = \fc(\x_{i,j}) \in \R^D$ is a
representation vector associated with the input region $(i,j)$ in
$\x$, i.e.\ corresponds to a specific receptive field or a patch in
the input image.  We then compute the output of the model based
completely on the coarse vectors, i.e.\ the coarse model's output
$\hc(\x) = \gtop(\fc(\x))$.



Next, we identify a few \emph{salient} input regions using an
attentional mechanism that exploits a \emph{saliency map} generated
using the coarse model's output.  The specific measure of saliency we
choose is based on the \emph{entropy} of the coarse model's output,
defined as:
\begin{equation}
  H = -\sum_{l=1}^C \vect{o}_c^{(l)}\log \vect{o}_c^{(l)},
\end{equation}
where $\vect{o}_c = \gtop(\fc(\x))$ is the vector output of the coarse
model and $C$ is the number of class labels.  The saliency $\M$ of an
input region position $(i,j)$ is given by the norm of the gradient of
the entropy $H$ with respect to the coarse vector $\vect{c}_{i,j}$:
\begin{align} \label{eq:saliency-map}
  \begin{split}
    M_{i,j} &= ||\nabla_{\vect{c}_{i,j}} H ||_2 \\
    &= \sqrt{\sum_{r=1}^D
      \left(\frac{\partial}{\partial\vect{c}_{i,j}^{(r)}}
        -\sum_{l=1}^C \vect{o}_c^{(l)}\log \vect{o}_c^{(l)}\right)^2},
  \end{split}
\end{align}
where $\M \in \R^{s_{1} \times s_{2}}$. The use of the entropy
gradient as a saliency measure encourages selecting input regions that
could affect the uncertainty in the model's predictions the most.  In
addition, computing the entropy of the output distribution does not
require observing the true label, hence the measure is available at
inference time. Note that computing all entries in matrix $\M$ can be
done using a single backward pass of backpropagation through the top
layers and is thus efficient and simple to implement.

Using the saliency map $\M$, we select a set of $k$ input region
positions with the highest saliency values.  We denote the selected
set of positions by
$\vect{I}^{s} \subseteq \left[1, s_1\right] \times \left[1,
  s_2\right]$, such that $| \vect{I}^s| = k$. We denote the set of
selected input regions by
$\vect{X}^{s}=\{\x_{i,j} \mid (i,j) \in \vect{I}^{s} \}$ where each
$\x_{i,j}$ is a patch in $\x$.  Next we apply the fine layers $\ff$
\emph{only on the selected patches} and obtain a small set of fine
representation vectors:
\begin{equation}
  \ff(\vect{X}^{s})=\{\vect{f}_{i,j} \mid (i,j) \in \vect{I}^{s} \},
\end{equation}
where $ \vect{f}_{i,j}=\ff(\x_{i,j})$.  This requires that
$\vect{f}_{i,j} \in \R^D$, i.e.\ the fine vectors have the same
dimensionality as the coarse vectors, allowing the model to use both
of them interchangeably.

We denote the representation resulting from combining vectors from
both $\fc(\x)$ and $\ff(\vect{X}^{s})$ as the \emph{refined
  representation} $f_{r}(\x)$.  We discuss in
Section~\ref{sec:experiments} different ways in which they can be
combined in practice.  Finally, the DCN output is obtained by feeding
the refined representation into the top layers, $\gtop(f_{r}(\x))$.
We denote the composition $\gtop \circ f_{r}$ by the \emph{refined
  model}.



\subsection{End-to-End Training} \label{sec:training-DCN}

In this section, we describe an end-to-end procedure for training the
DCN model that leverages our attention mechanism to learn $\ff$ and
$\fc$ jointly.  We emphasize, however, that DCN modules can be trained
independently, by training a coarse and a fine model independently and
combining them only at test-time using our attention based inference.
In Section~\ref{sec:svhn-exp} we show an example of how this modular
training can be used for transfer learning.

In the context of image classification, suppose we have a training set
$\D=\{(\xsi,\ysi);i=1\dots m\}$, where each $\xsi \in \R^{h\times w}$
is an image, and $\ysi \in \{1,\dots,C\}$ is its corresponding
label. We denote the parameters of the coarse, fine and top layers by
$\theta_c$, $\theta_f$, and $\theta_t$ respectively.  We learn all of
these parameters (denoted as $\theta$) by minimizing the cross-entropy
objective function (which is equivalent to maximizing the
log-likelihood of the correct labels):
\begin{equation} \label{eq:cross-entropy} J = -\sum_{i=1}^{m} \log p
  \left( \ysi \mid \xsi; \theta \right),
\end{equation}
where $p(\cdot \mid \xsi ; \theta)=\gtop(f_{r}(\xsi))$ is the
conditional multinomial distribution defined over the $C$ labels given
by the refined model (Figure \ref{figure:DCN}). Gradients are computed
by standard back-propagation through the refined model, i.e.\
propagating gradients at each position into either the coarse or fine
features, depending on which was used.

An important aspect of the DCN model is that the final prediction is
based on combining representations from two different sets of layers,
namely the coarse layers $\fc$ and the fine layers $\ff$. Intuitively,
we would like those representations to have close values such that
they can be interchangeable. This is important for two reasons. First,
we expect the top layers to have more success in correctly classifying
the input if the transition from coarse to fine representations is
smooth. The second is that, since the saliency map is based on the
gradient \emph{at the coarse representation values} and since the
gradient is a local measure of variation, it is less likely to reflect
the benefit of using the fine features if the latter is very different
from the former.

To encourage similarity between the coarse and fine representations
while training, we use a hint-based training approach inspired by
\citet{romero2014fitnets}.  Specifically, we add an additional term to
the training objective that minimizes the squared distance between
coarse and fine representations:
\begin{equation} \label{eq:hints} \sum_{\x_{i,j} \in \vect{X}^s}
  \|\fc(\x_{i,j})-\ff(\x_{i,j}) \|^2_2.
\end{equation}
There are two important points to note here. First, we use this term
to optimize \emph{only the coarse layers} $\theta_c$. That is, we
encourage the coarse layers to mimic the fine ones, and let the fine
layers focus only on the signal coming from the top layers.  Secondly,
computing the above \emph{hint} objective over representations at all
positions would be as expensive as computing the full fine model;
therefore, we encourage in this term similarity only over the selected
salient patches.

\section{Related Work}

This work can be classified as a conditional computation approach.
The goal of conditional computation, as put forward by
\citet{bengio2013deep}, is to train very large models for the same
computational cost as smaller ones, by avoiding certain computation
paths depending on the input. There have been several contributions in
this direction. \citet{bengio2013estimating} use stochastic neurons as
gating units that activate specific parts of a neural network. Our
approach, on the other hand, uses a hard-attention mechanism that
helps the model to focus its computationally expensive paths only on
important input regions, which helps in both scaling to larger
effective models and larger input sizes.

Several recent contributions use attention mechanisms to capture
visual structure with biologically inspired, foveation-like methods,
e.g.\ \cite{larochelle2010learning, denil2012learning,
  ranzato2014learning, mnih2014recurrent, ba2014multiple,
  gregor2015draw}.  In \citet{mnih2014recurrent, ba2014multiple}, a
learned sequential attention model is used to make a hard decision as
to where to look in the image, i.e.\ which region of the image is
considered in each time step.  This so-called ``hard-attention''
mechanism can reduce computation for inference. The attention
mechanism is trained by reinforcement learning using policy search.
In practice, this approach can be computationally expensive during
training, due to the need to sample multiple interaction sequences
with the environment.  On the other hand, the DRAW model
\cite{gregor2015draw} uses a ``soft-attention'' mechanism that is
fully differentiable, but requires processing the whole input at each
time step. Our approach provides a simpler hard-attention mechanism
with computational advantages in both inference and learning.

The saliency measure employed by DCN's attention mechanism is related
to pixel-wise saliency measures used in visualizing neural
networks~\cite{simonyan2013deep}. These measures, however, are based
on the gradient of the classification loss, which is not applicable at
test-time.
Moreover, our saliency measure is defined over contiguous regions of
the input rather than on individual pixels. It is also task-dependent,
as a result of defining it using a coarse model trained on the same
task.

Other works such as matrix factorization \citep{Jaderberg14,Denton14}
and quantization schemes \citep{Chen10,Jegou11,Gong14} take the same
computational shortcuts for all instances of the data.  In contrast,
the shortcuts taken by DCN specialize to the input, avoiding costly
computation except where needed.  However, the two approaches are
orthogonal and could be combined to yield further savings.

Our use of a regression cost for enforcing representations to be
similar is related to previous work on model compression
\cite{bucilua2006model, hinton2015distilling, romero2014fitnets}. The
goal of model compression is to train a small model (which is faster
in deployment) to imitate a much larger model or an ensemble of
models. Furthermore, \citet{romero2014fitnets} have shown that middle
layer hints can improve learning in deep and thin neural networks.
Our DCN model can be interpreted as performing model compression on
the fly, without the need to train a large model up front.

\section{Experiments} \label{sec:experiments}

In this section, we present an experimental evaluation of the proposed
DCN model.  To validate the effectiveness of our approach, we first
investigate the Cluttered MNIST dataset~\cite{mnih2014recurrent}.  We
then apply our model in a transfer learning setting to a real-world
object recognition task using the Street View House Numbers (SVHN)
dataset~\cite{netzer2011reading}.

\subsection{Cluttered MNIST} \label{sec:mnist-exp}

We use the $100\times 100$ Cluttered MNIST digit classification
dataset \cite{mnih2014recurrent}. Each image in this dataset is a
hand-written MNIST digit located randomly on a $100\times 100$ black
canvas and cluttered with digit-like fragments. Therefore, the dataset
has the same size of MNIST: 60000 images for training and 10000 for
testing.

\subsubsection{Model Specification}
In this experiment we train a DCN model end-to-end, where we learn
coarse and fine layers jointly. We use 2 convolutional layers as
coarse layers, 5 convolutional layers as fine layers and one
convolutional layer followed by global max pooling and a softmax as
the top layers. Details of their architectures can be found in the
Appendix~\ref{sec:mnist-arch}.  The coarse and fine layers produce
feature maps, i.e.\ feature vectors each with a specific spatial
location.  The set of selected patches $\vect{X}^s$ is composed of
eight patches of size $14\times 14$ pixels.  We use here a refined
representation of the full input $f_{r}(\x)$ in which fine feature
vectors are swapped in place of coarse ones:

\begin{eqnarray} f_r(\x) &= \{ \vect{r}_{i,j} \mid (i,j) \in \left[1,
    s_1\right] \times
                           \left[1, s_2\right] \}\\
  \vect{r}_{i,j} &= \begin{cases}
    \ff(\x_{i,j}),& \text{if } \x_{i,j} \in \vect{X}^s\\
    \fc(\x_{i,j}),& \text{otherwise}.
  \end{cases}
\end{eqnarray}

\subsubsection{Baselines}

We use as baselines for our evaluation the coarse model (top layers
applied only on coarse representations), the fine model (top layers
applied only on fine representations), and we compare with previous
attention-based models RAM~\cite{mnih2014recurrent} and
DRAW~\cite{gregor2015draw}.

\subsubsection{Empirical Evaluation}

\begin{table}[t]\ \centering
\small
  \begin{tabular}{ll}
    \toprule
    Model       & Test Error \\
    \midrule
    RAM         & 8.11\%     \\
    DRAW        & 3.36\%     \\
    \hline
    Coarse Model & 3.69\%     \\
    Fine Model   & 1.70\%     \\
    DCN w/o hints        & 1.71\% \\
    DCN with hints        & \textbf{1.39}\% \\
    \bottomrule
  \end{tabular}
  \caption{Results on Cluttered MNIST}
  \label{tab:cluttered-results}
\end{table}

\begin{figure}[t]
  \centering
  \includegraphics[width=\columnwidth]{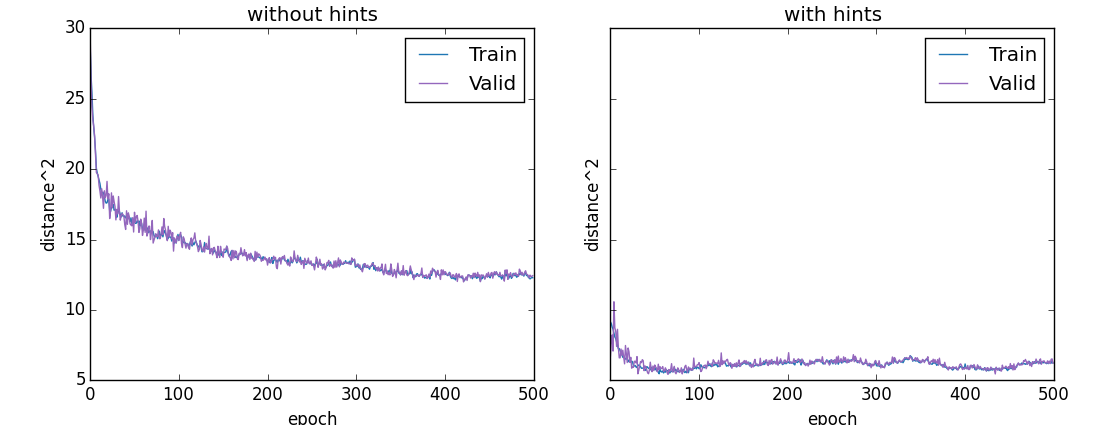}
  \caption{The effect of using the hints objective. We show the
    squared distance between coarse and fine features over salient
    regions during training in two cases: with and without using the
    hints objective.  We observe that this regularizer helps in
    minimizing the distance and improves the model's generalization.}
  \label{figure:hints-effect}
\end{figure}

\begin{figure}[t]
  \center \subfigure{
    \label{figure:patches}
    \includegraphics[width=0.4\columnwidth]{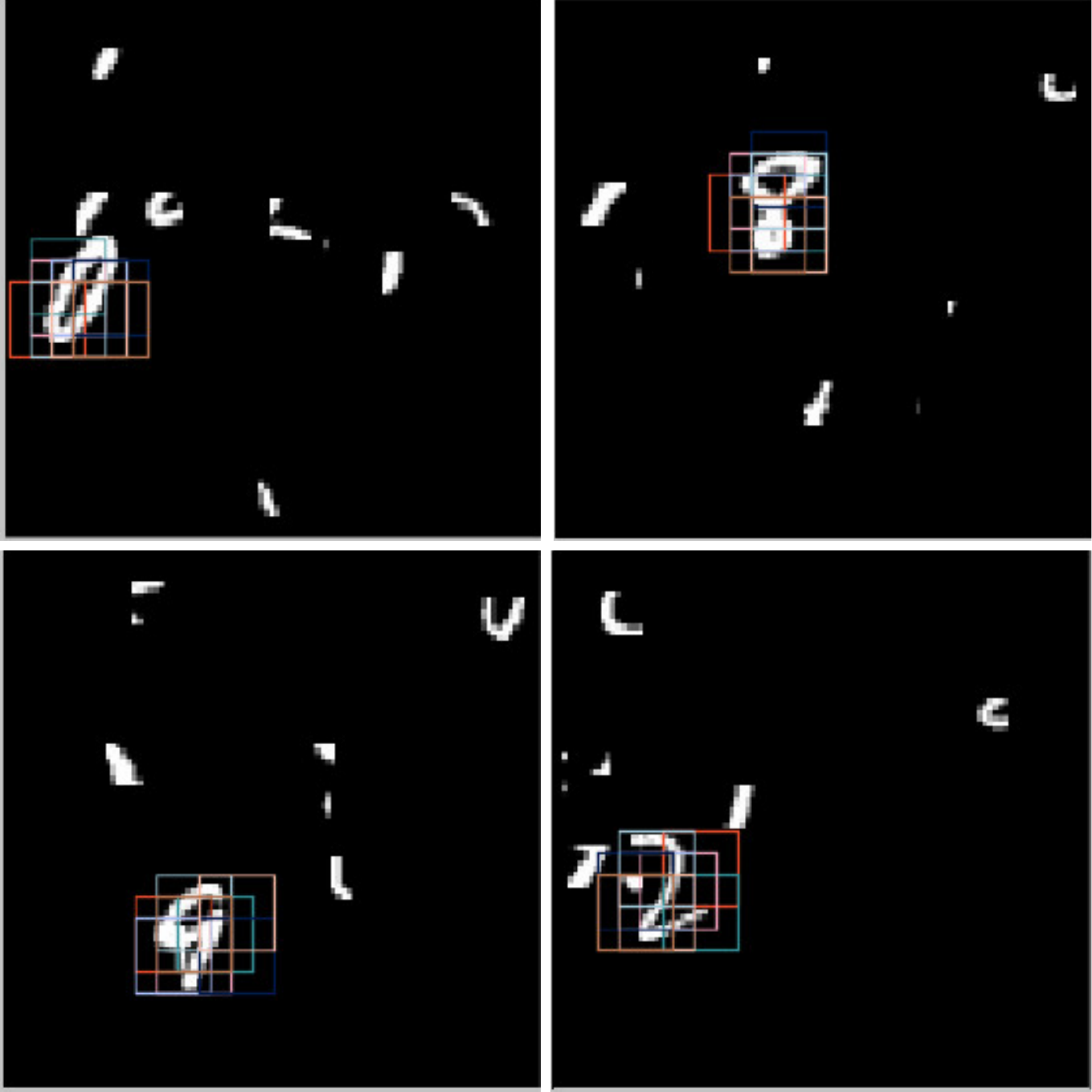}
  } \hspace{2mm} \subfigure{
    \label{figure:k_vs_error}
    \includegraphics[width=.5\columnwidth]{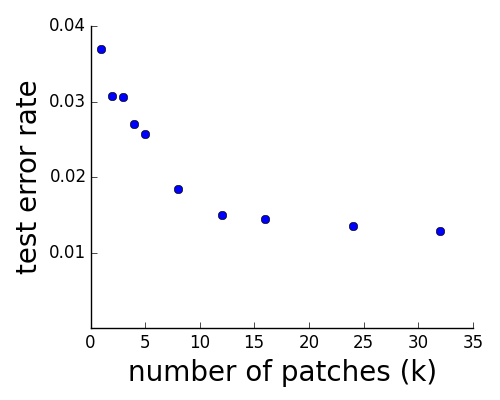}
  }
  \caption{Patch selection on Cluttered MNIST. 
(\textit{left}) Sample of selected patches. 
(\textit{right}) Test error vs. number of patches: taking
   more patches yields lower error, but with diminishing returns.}
  \label{figure:DCNtradeoff}
\end{figure}

Results of our experiments are shown in Table
\ref{tab:cluttered-results}.  We get our best DCN result when we add
the hint term in Eq.~\eqref{eq:hints} in the training objective, which
we observe to have a regularization effect on DCN.  We can see that
the DCN model performs significantly better than the previous
state-of-the-art result achieved by RAM and DRAW models. It also
outperforms the fine model, which is a result of being able to focus
only on the digit and ignore clutter.  In
Figure~\ref{figure:hints-effect} we explore more the effect of the
hint objective during training, and confirm that it can indeed
minimize the squared distance between coarse and fine representations.
To show how the attention mechanism of the DCN model can help it focus
on the digit, we plot in Figure~\ref{figure:patches} the patches it
finds in some images from the validation set, after only 9 epochs of
training.

The DCN model is also more computationally efficient.  A forward pass
of the fine model requires the computation of the fine layers
representations on whole inputs and a forward pass of the top layers
leading to 84.5M multiplications.  On the other hand, DCN applies only
the coarse layers on the whole input.  It also requires the
computation of the fine representations for 8 input patches and a
forward pass of the top layers.  The attention mechanism of the DCN
model requires an additional forward and backward pass through the top
layers which leads to approximately $27.7$M multiplications in total.
As a result, the DCN model here has 3 times fewer multiplications than
the fine model. In practice we observed a time speed-up by a factor of
about 2.9. Figure~\ref{figure:k_vs_error} shows how the test error behaves when
we increase the number of patches.  While taking additional patches
improves accuracy, the marginal improvement becomes insignificant
beyond 10 or so patches.  The number of patches effectively controls a
trade-off between accuracy and computational cost.


\subsection{SVHN} \label{sec:svhn-exp} We tackle in this section a
more challenging task of transcribing multi-digit sequences from
natural images using the Street View House Numbers (SVHN) dataset
~\cite{netzer2011reading}. SVHN is composed of real-world pictures
containing house numbers and taken from house fronts.  The task is to
recognize the full digit sequence corresponding to a house number,
which can be of length 1 to 5 digits. The dataset has three subsets:
train (33k), extra (202k) and test (13k).  In the following, we
trained our models on 230k images from both the train and extra
subsets, where we take a 5k random sample as a validation set for
choosing hyper-parameters.

The typical experimental setting in previous literature, e.g.\
\cite{goodfellow2013multi, ba2014multiple, jaderberg2015spatial}, uses
the location of digit bounding boxes as extra information.  Input
images are generally cropped, such that digit sequences are centred
and most of the background and clutter information is pruned.  We
argue that our DCN model can deal effectively with real-world noisy
images having large portions of clutter or background information.  To
demonstrate this ability, we investigate a more general problem
setting where the images are uncropped and the digits locations are
unknown.  We apply our models on SVHN images \emph{in their original
  sizes and we do not use any extra bounding box information}.
\footnote{The only pre-processing we perform on the data is converting
  images to grayscale.}

\begin{figure}[t]
  \centering
  \includegraphics[width=.4\columnwidth]{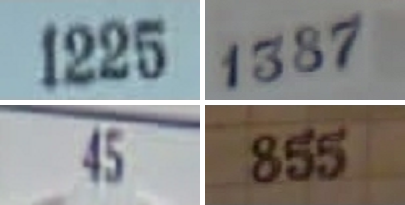}
  \hspace{2mm}
  \includegraphics[width=.4\columnwidth]{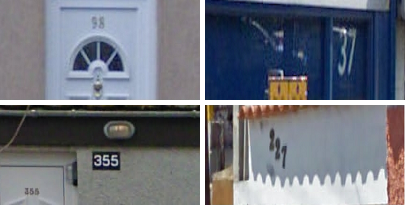}
  \caption{The 4 left images are samples from the extra subset, and
    the 4 right images are samples from the test subset.  We notice
    that extra images are well-centred and have much less background
    compared to test images.}
  \label{figure:svhn-images}
\end{figure}

An important property of the SVHN dataset is the large discrepancy
between the train/extra sets and the test set. Most of the extra
subset images (which dominate the training data) have their digits
well-centred with little cluttered background, while test images have
more variety in terms of digit location and background clutter.
Figure~\ref{figure:svhn-images} shows samples of these images.  We can
tackle this training/test dataset discrepancy by training a DCN model
in a transfer learning setting.  We train the coarse and fine layers
of the DCN independently on the training images that have little
background-clutter, and then combine them using our attention
mechanism, which does not require explicit training, to decide on
which subsets of the input to apply the fine layers.


\subsubsection{Multi-Digit Recognition Model} \label{sec:svhn-output}

We follow the model proposed in~\cite{goodfellow2013multi} for
learning a probabilistic model of the digit sequence given an input
image $\x$.  The output sequence $\vect{S}$ is defined using a
collection of $N$ random variables, $S_1, \dots, S_N$, representing
the elements of the sequence and an extra random variable $S_0$
representing its length.  The probability of a given sequence
$\vect{s}=\{ s_1, \dots, s_n\}$ is given by:
\begin{equation}\label{eq:svhnout}
  p(\vect{S}=\vect{s} \mid \x) = p(S_0=n \mid \x) \prod_{i=1}^n p(S_i=s_i \mid \x),
\end{equation}
where $p(S_0=n \mid \mathbf{x})$ is the conditional distribution of
the sequence length and $p(S_i=s_i \mid \mathbf{x})$ is the
conditional distribution of the $i$-th digit in the sequence.  In
particular, our model on SVHN has 6 softmaxes: 1 for the length of the
sequence (from $1$ to $5$), and 5 for the identity of each digit or a
null character if no digit is present (11 categories).

\subsubsection{Model Specification}

The coarse and fine bottom layers, $\fc$ and $\ff$, are
fully-convolutional, composed of respectively $7$ and $11$ layers.
The representation, produced by either the fine or coarse layers, is a
\emph{probability map}, which is a collection of independent
full-sequence prediction vectors, each vector corresponding to a
specific region of the input.  We denote the prediction for the $i$-th
output at position $(j,k)$ by $p^{(j,k)}(S_{i} \mid \x)$.

The top layer $g$ is composed of one global average pooling layer
which combines predictions from various spatial locations to produce
the final prediction $p(\vect{S} \mid \x)$.



Since we have multiple outputs in this task, we modify the saliency
measure used by the DCN's attention mechanism to be the sum of the
entropy of the 5 digit softmaxes:
\begin{equation}
  H = - \sum_{i=1}^5 \sum_{j=1}^{11} p(S_i=s_j \mid \x)\log p(S_i=s_j \mid \x).
\end{equation}

When constructing the saliency, instead of using the gradient with
respect to the probability map, we use the gradient with respect to
the feature map below it. This is necessary to avoid identical
gradients as $g$, the top function, is composed by only one average
pooling.

We also use a refined model that computes its output by applying the
pooling top layer $g$ only on the $k$ independent predictions from
fine layers, ignoring the coarse layers. We have found empirically
that this results in a better model, and suspect that otherwise the
predictions from the salient regions are drowned out by the noisy
predictions from uninformative regions.

We train the coarse and fine layers of DCN independently in this
experiment, minimizing $\log p(\vect{S} \mid \x)$ using SGD. For the
purposes of training only, we resize images to $64 \times 128$.
Details on the coarse and fine architectures are found in
Appendix~\ref{sec:svhn-arch}.



\subsubsection{Baselines}

As mentioned in the previous section, each of the coarse
representation vectors in this experiment corresponds to multi-digit
recognition probabilities computed at a given region, which the top
layer $g$ simply averages to obtain the baseline coarse model:
\begin{equation}
  p(S_i \mid \x) = \frac{1}{d_1\times d_2}\sum_{j,k} p^{(j,k)}(S_{i} \mid \vect{x}).
\end{equation}
The baseline fine model is defined similarly.

As an additional baseline, we consider a ``soft-attention'' coarse
model, which takes the coarse representation vectors over all input
regions, but uses a top layer that performs a weighted average of the
resulting location-specific predictions.  We leverage the entropy to
define a weighting scheme which emphasizes important locations:
\begin{equation} \label{eq:weighted-sum} p(S_i \mid \x) = \sum_{j,k}
  w_{i,j,k} p^{(j,k)}(S_{i} \mid \vect{x}).
\end{equation}
The weight $w_{i,j,k}$ is defined as the \emph{normalized inverse
  entropy} of the $i$-th prediction by the $(j,k)$-th vector, i.e.\ :
\begin{equation}
  w_{i,j,k} = \sum_{q,r} \frac{H_{i,j,k}^{-1}}{H_{i,q,r}^{-1}},
\end{equation}
where $H_{i,j,k}$ is defined as:
\begin{equation}
  H_{i,j,k} = -\sum_{l=1}^C p^{j,k}(S_{i}=s_l \mid \vect{x}) \log p^{j,k}(S_{i}=s_l \mid \vect{x}),
\end{equation}
and $C$ is either $5$ for $S_0$ or $11$ for all other $S_i$.  As we'll
see, this weighting improves the coarse model's performance in our
SVHN experiments. We incorporate this weighting in DCN to aggregate
predictions from the salient regions.

To address scale variations in the data, we extend all models to
multi-scale by processing each image several times at multiple
resolutions.  Predictions made at different scales are considered
independent and averaged to produce the final prediction.

It is worth noting that all previous literature on SVHN dealt with a
simpler task where images are cropped and resized. In this experiment
we deal with a more general setting, and our results cannot be
directly compared with these results.

\begin{table}[t]\ \centering
  \small
  \begin{tabular}{ll}
    \toprule
    Model       & Test Error \\
    \midrule
    Coarse model, 1 scale & 40.6\% \\
    Coarse model, 2 scales & 40.0\% \\
    Coarse model, 3 scales & 40.0\% \\
    \midrule
    Fine model, 1 scale & 25.2\% \\
    Fine model, 2 scales & 23.7\% \\
    Fine model, 3 scales & 23.3\% \\
    \midrule
    Soft-attention, 1 scale & 31.4\% \\
    Soft-attention, 2 scales & 31.1\% \\
    Soft-attention, 3 scales & 30.8\% \\
    \midrule
    DCN, 6 patches, 1 scale & 20.0\% \\
    DCN, 6 patches, 2 scales & 18.2\% \\
    DCN, 9 patches, 3 scales & 16.6\% \\
    \bottomrule
  \end{tabular}
  \caption{Results on SVHN dataset without using bounding box
    information.}%
  \label{tab:svhn-results}
\end{table}

\begin{figure}[t]
  \center
  \includegraphics[width=.65\columnwidth]{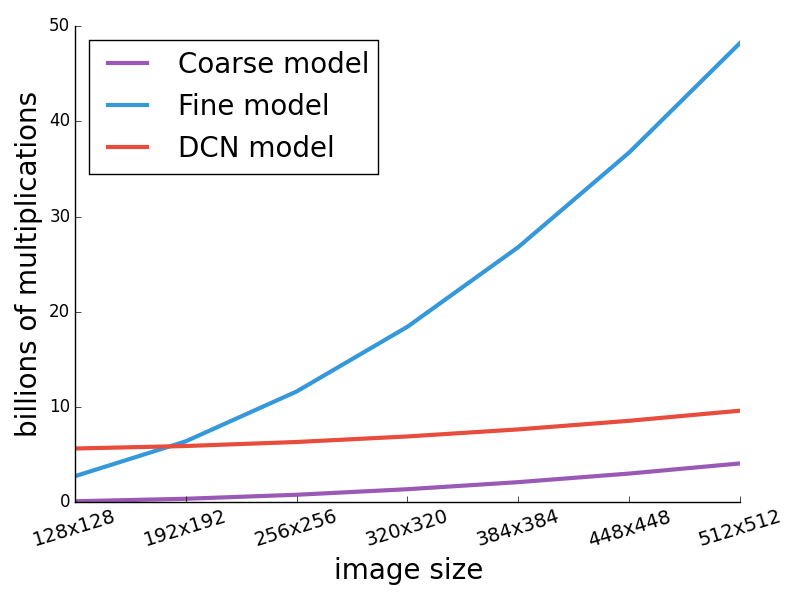}
  \caption{Number of multiplications in the Coarse, Fine and DCN
    models given different image input sizes.}
  \label{figure:dcncomputation}
\end{figure}

\begin{figure}[t]
  \center
  \includegraphics[width=.65\columnwidth]{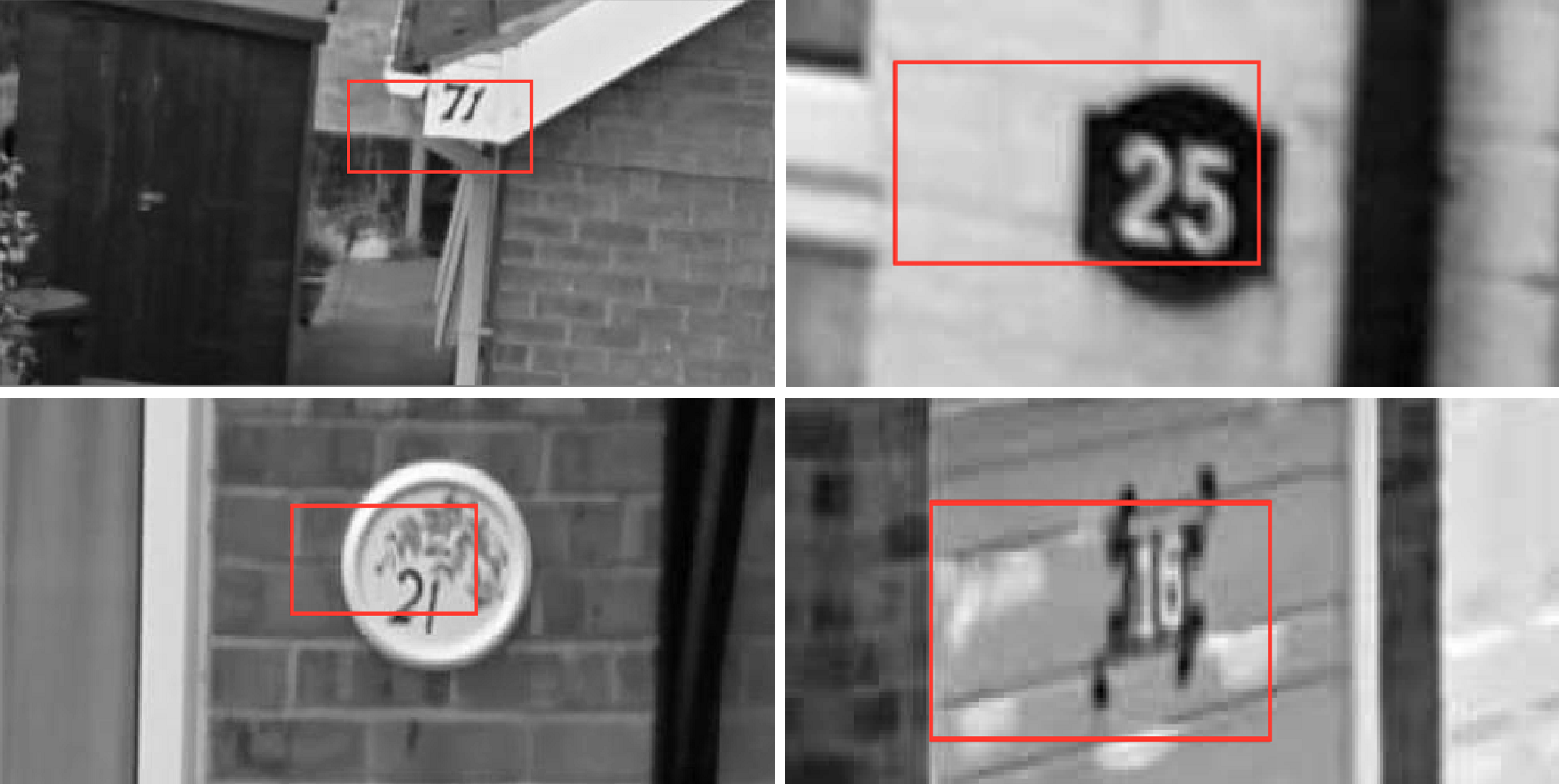}
  \caption{A sample of the selected patches in SVHN images.  The
    images are processed by the DCN inference procedure in their
    original sizes. They are resized here for illustration purposes.}
  \label{fig:svhn_patches}
\end{figure}

\subsubsection{Empirical Evaluation}

Table~\ref{tab:svhn-results} shows results of our experiment on SVHN.
The coarse model has an error rate of $40.6\%$, while by using our
proposed soft-attention mechanism, we decrease the error rate to
$31.4\%$. This confirms that the entropy is a good measure for
identifying important regions when task-relevant information is not
uniformly distributed across input data.

The fine model, on the other hand, achieves a better error rate of
$25.2\%$, but is more computationally expensive.  Our DCN model, which
selects only 6 regions on which to apply the high-capacity fine
layers, achieves an error rate of $20.0\%$.  The DCN model can
therefore outperform, in terms of classification accuracy, the other
baselines.  This verifies our assumption that by applying high
capacity sub-networks only on the input’s most informative regions, we
are able to obtain high classification performance.
Figure~\ref{fig:svhn_patches} shows a sample of the selected patches
by our attention mechanism.

An additional decrease of the test errors can be obtained by
increasing the number of processed scales. In the DCN model, taking 3
patches at 2 scales (original and 0.75 scales), leads to $18.2\%$
error, while taking 3 patches at 3 scales (original, 0.75 and 0.5
scales) leads to an error rate of $16.6\%$. Our DCN model can reach
its best performance of $11.6\%$ by taking all possible patches at 3
scales, but it does not offer any computational benefit over the fine
model.

We also investigate the computational benefits of the DCN approach as
the dimensions of the input data increase.
Table~\ref{figure:dcncomputation} reports the number of
multiplications the fine model, coarse model and the DCN model
require, given different input sizes.
We also verify the actual computational time of these models by taking
the largest 100 images in the SVHN test set, and computing the average
inference time taken by all the models.~\footnote{We evaluate all
  models on an NVIDIA Titan Black GPU card.} The smallest of these
images has a size of $363 \times 735$ pixels, while the largest has a
size of $442 \times 1083$ pixels.  On average, the coarse and the
soft-attention models take $8.6$ milliseconds, while the fine model
takes $62.6$ milliseconds.  On the largest 100 SVHN test images, the
DCN requires on average $10.8$ milliseconds for inference.

\section{Conclusions}

We have presented the DCN model, which is a novel approach for
conditional computation.  We have shown that using our visual
attention mechanism, our network can adaptively assign its capacity
across different portions of the input data, focusing on important
regions of the input. Our model achieved state-of-the-art performance
on the Cluttered MNIST digit classification task, and provided
computational benefits over traditional convolutional network
architectures.  We have also validated our model in a transfer
learning setting using the SVHN dataset, where we tackled the
multi-digit recognition problem without using any a~priori information
on the digits' location.  We have shown that our model outperforms
other baselines, yet remains tractable for inputs with large spatial
dimensions.

\section{Appendix}
\subsection{Cluttered MNIST Experiment Details} \label{sec:mnist-arch}
\begin{itemize}
\item Coarse layers: 2 convolutional layers, with $7\times 7$ and
  $3\times 3$ filter sizes, 12 and 24 filters, respectively, and a
  $2\times 2$ stride. Each feature in the coarse feature maps covers a
  patch of size $11\times 11$ pixels, which we extend by $3$ pixels in
  each side to give the fine layers more context. The size of the
  coarse feature map is $23\times 23$.
\item Fine layers: 5 convolutional layers, each with $3\times 3$
  filter sizes, $1\times 1$ strides, and 24 filters. We apply
  $2\times 2$ pooling with $2\times 2$ stride after the second and
  fourth layers. We also use $1\times 1$ zero padding in all layers
  except for the first and last layers. This architecture was chosen
  so that it maps a $14\times 14$ patch into one spatial location.
\item Top layers: one convolutional layer with $4\times 4$ filter
  size, $2\times 2$ stride and 96 filters, followed by global max
  pooling. The result is fed into a 10-output softmax layer.
\end{itemize}

We use rectifier non-linearities in all layers. We use Batch
Normalization \cite{ioffe2015batch} and Adam \cite{kingma2014adam} for
training our models. In DCN we train the coarse layers with a convex
combination of cross entropy objective and hints.

\subsection{SVHN Experiment Details} \label{sec:svhn-arch}
\begin{itemize}
\item Coarse layers: the model is fully convolutional with 7
  convolutional layers.  First three layers have 24, 48, 128 filters
  respectively with size $5\times 5$ and stride $2\times 2$. Layer 4
  has 192 filters with $4\times 5$ and stride $1\times 2$. Layer 5 has
  192 filters with size $1\times 4$. Finally, the last two layers are
  $1\times 1$ convolutions with 1024 filters. We use stride of
  $1\times 1$ in the last 3 layers and do not use zero padding in any
  of the coarse layers.  The corresponding patch size here is
  $54\times 110$.

\item Fine layers: 11 convolutional layers.  The first 5 convolutional
  layers have 48, 64, 128, 160 and 192 filters respectively, with size
  $5\times 5$ and zero-padding. After layers 1, 3, and 5 we use
  $2 \times 2$ max pooling with stride $2\times 2$.  The following
  layers have $3\times 3$ convolution with 192 filters. The 3 last
  layers are $1\times 1$ convolution with 1024 hidden units.
\end{itemize}

Here we use SGD with momentum and exponential learning rate decay.
While training, we take $54\times 110$ random crop from images, and we
use 0.2 dropout on convolutional layers and 0.5 dropout on fully
connected layers.

\section*{Acknowledgements}
The authors would like to acknowledge the support of the following
organizations for research funding and computing support: Nuance
Foundation, Compute Canada and Calcul Qu\'ebec.  We would like to
thank the developers of Theano \cite{bergstra+all-Theano-NIPS2011,
  Bastien-Theano-2012} and Blocks/Fuel \cite{Van+al-arxiv-2015} for
developing such powerful tools for scientific computing, and our
reviewers for their useful comments.







\bibliography{local,aigaion} \bibliographystyle{icml2016}

\end{document}